%% file: main.tex
\documentclass[10pt,twocolumn,letterpaper]{article}

\usepackage[pagenumbers]{cvpr} 

\input{figures/include}

\definecolor{cvprblue}{rgb}{0.21,0.49,0.74}
\usepackage[pagebackref,breaklinks,colorlinks,allcolors=cvprblue]{hyperref}
\usepackage{arydshln}
\usepackage{graphicx}
\usepackage{float}

\def\paperID{3356} 
\def\confName{CVPR}
\def\confYear{2025}

\title{\includegraphics[height=25pt]{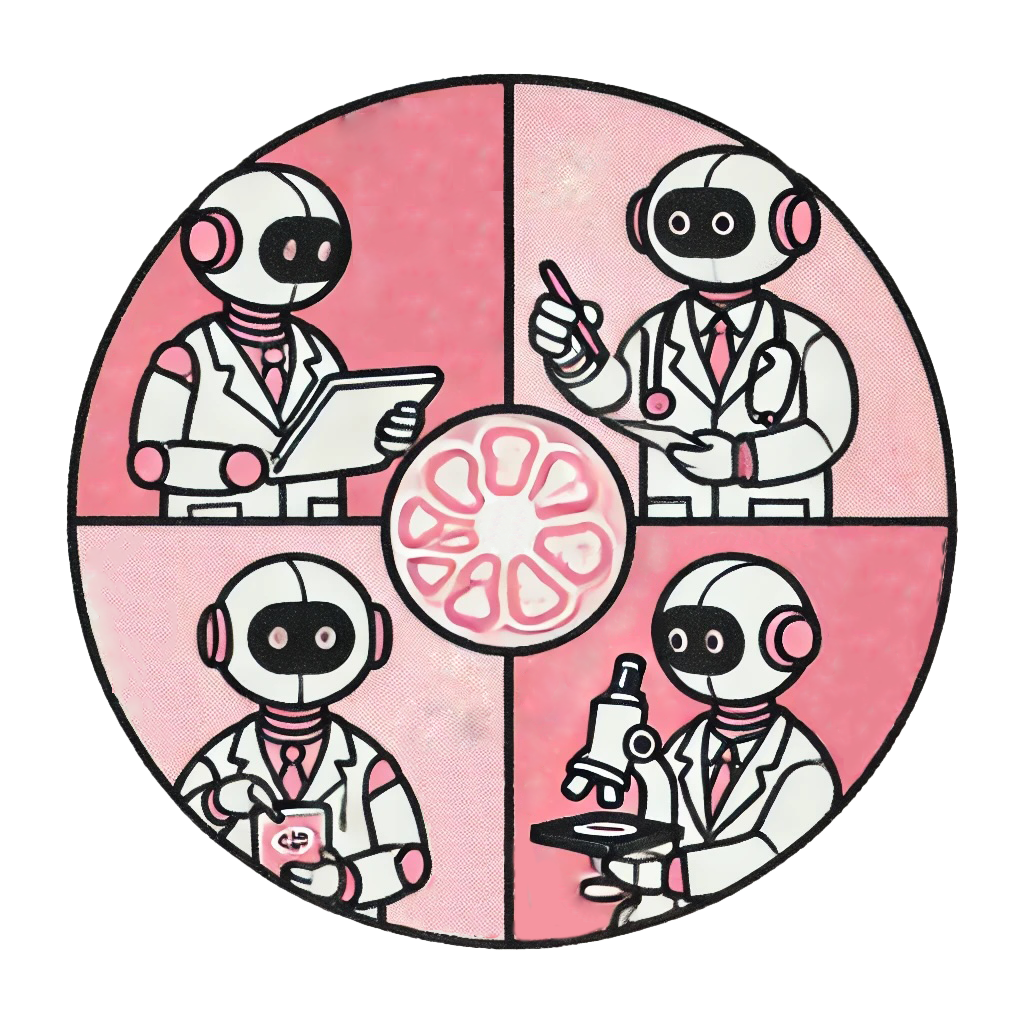} PathFinder: A Multi-Modal Multi-Agent System \\for Medical Diagnostic Decision-Making Applied to Histopathology}

\author{ \bf Fatemeh Ghezloo $^{1,2*}$\qquad
\bf Mehmet Saygin Seyfioglu  $^{1*}$ \qquad
\bf Rustin Soraki $^{1*}$ \qquad \\
\bf Wisdom O. Ikezogwo $^{1*}$ \qquad
\bf Beibin Li $^{2*}$ \qquad
\bf Tejoram Vivekanandan $^{1}$ \qquad \\
\bf Joann G. Elmore $^{3}$ \qquad
\bf Ranjay Krishna $^{1,4}$  \qquad
\bf Linda Shapiro $^{1}$  \qquad\\
$^{1}$ University of Washington \qquad
$^{2}$ Microsoft Research \\
$^{3}$ David Geffen School of Medicine, UCLA \qquad
$^{4}$ Allen Institute for AI (AI2) \\
}

\begin{document}

\twocolumn[{%
\renewcommand\twocolumn[1][]{#1}%
\maketitle

\begin{center}
    \centering
    \captionsetup{type=figure}
    \includegraphics[width=1\textwidth, height=4.5cm]{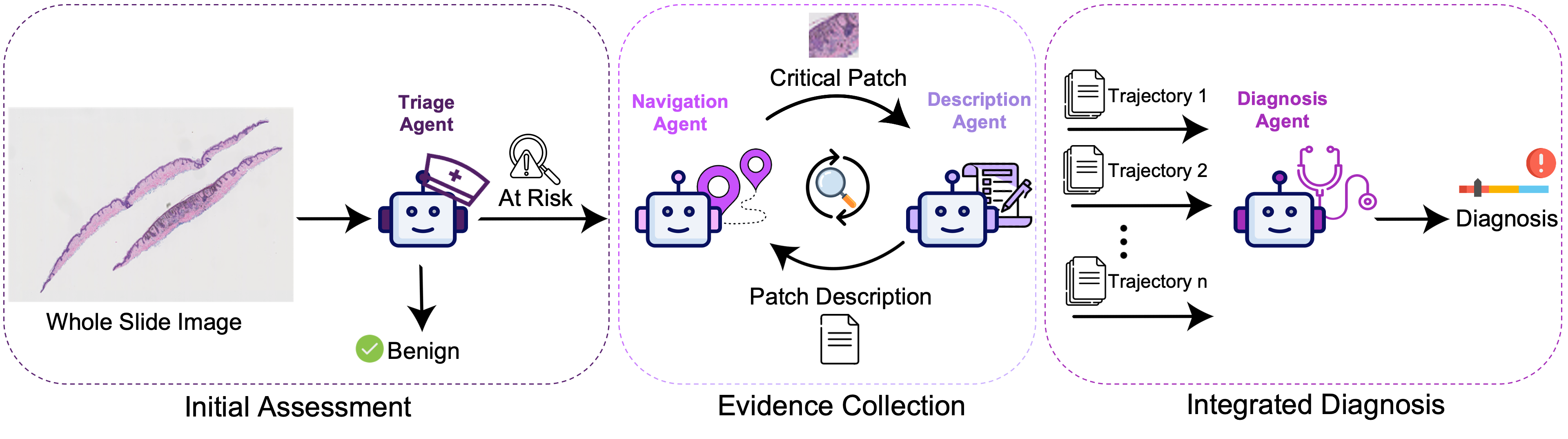}
   \caption{We propose \textbf{PathFinder}, a system capable of navigating image patches in a whole slide image, describe each patch to collect evidence, and produce a diagnosis. Pathfinder's process is interpretable and reminiscent of pathologists. Our system consists of multiple steps carried out by multi-modal agents: 1) Initial Assessment by Triage Agent; 2) Evidence Collection by Navigation and Description Agents; and 3) Integrated Diagnosis by Diagnosis Agent.}
\label{fig:nav-pipeline-simple}
\end{center}%
}]

\maketitle
\input{sec/0_abstract}

\input{sec/1_intro}
\input{sec/2_related_work}
\input{sec/3_datasets}

\input{sec/3_methods}

\input{sec/4_Experiments}
\input{sec/5_Discussion}

\noindent\textbf{Acknowledgements.} We thank Dr. Oliver Chang and Dr. Kristin Shaw for their participation in our clinical evaluation. We also acknowledge Microsoft for providing OpenAI credits, Department of Defense (W81XWH-20-1-0798),  Na
tional Cancer Institute (U01CA231782, and R01CA200690), and partial funding through a Population Health Initiative at University of Washington.

{
    \small
    \bibliographystyle{ieeenat_fullname}
    \bibliography{main}
}

\input{sec/X_suppl}

\end{document}

%% file: figures/include.tex
\usepackage{tikz}
\usepackage{pgfplots, pgfplotstable}
\pgfplotsset{compat=newest}
\usepgfplotslibrary{external}
\usetikzlibrary{fit, backgrounds}
\usetikzlibrary{arrows,positioning}
\usetikzlibrary{arrows.meta, positioning}
\usetikzlibrary{ calc, matrix, patterns, patterns.meta }
\usetikzlibrary{shapes,arrows,positioning, fit, tikzmark}
\usetikzlibrary{arrows}
\usepgfplotslibrary{statistics}
\usetikzlibrary{decorations.pathreplacing}
\usepgfplotslibrary{fillbetween}
\definecolor{msdarkblue}{RGB}{36,58,94}
\definecolor{msblue}{RGB}{0,120,215}
\definecolor{msgreen}{RGB}{16,124,16}
\definecolor{msred}{RGB}{216,59,1}
\definecolor{msgray}{HTML}{DFDFDF}

\usetikzlibrary{arrows}
\usepgfplotslibrary{statistics}
\usetikzlibrary{decorations.pathreplacing}
\usepgfplotslibrary{fillbetween}
\definecolor{msdarkblue}{RGB}{36,58,94}
\definecolor{msblue}{RGB}{0,120,215}
\definecolor{msgreen}{RGB}{16,124,16}
\definecolor{msred}{RGB}{216,59,1}
\definecolor{msgray}{HTML}{DFDFDF}
\pgfplotsset{compat=1.6}

%% file: sec/0_abstract.tex
\begin{abstract}

Diagnosing diseases through histopathology whole slide images (WSIs) is fundamental in modern pathology but is challenged by the gigapixel scale and complexity of WSIs. Trained histopathologists overcome this challenge by navigating the WSI, looking for relevant patches, taking notes, and compiling them to produce a final holistic diagnostic. Traditional AI approaches, such as multiple instance learning and transformer-based models, fail short of such a holistic, iterative, multi-scale diagnostic procedure, limiting their adoption in the real-world. We introduce PathFinder, a multi-modal, multi-agent framework that emulates the decision-making process of expert pathologists. PathFinder integrates four AI agents—the Triage Agent, Navigation Agent, Description Agent, and Diagnosis Agent—that collaboratively navigate WSIs, gather evidence, and provide comprehensive diagnoses with natural language explanations. The Triage Agent classifies the WSI as benign or risky; if risky, the Navigation and Description Agents iteratively focus on significant regions, generating importance maps and descriptive insights of sampled patches. Finally, the Diagnosis Agent synthesizes the findings to determine the patient's diagnostic classification. Our Experiments show that PathFinder outperforms state-of-the-art methods in skin melanoma diagnosis by 8\% while offering inherent explainability through natural language descriptions of diagnostically relevant patches. Qualitative analysis by pathologists shows that the Description Agent’s outputs are of high quality and comparable to GPT-4o. PathFinder is also the first AI-based system to surpass the average performance of  pathologists in this challenging melanoma classification task by 9\%, setting a new record for efficient, accurate, and interpretable AI-assisted diagnostics in pathology.  Data, demo, code and models are available at \href{https://pathfinder-dx.github.io/}{https://pathfinder-dx.github.io/}.

\vspace{-5pt}
\end{abstract}

%% file: sec/1_intro.tex
\section{Introduction}
\label{sec:intro}

Medical diagnosis of histopathology through the examination of whole slide images (WSIs) is a cornerstone of modern pathology. WSIs are high-resolution, digitally scanned histopathology cases, providing an extensive view of tissue architecture and cellular detail. 
Pathologists \textit{navigate} these gigapixel-scale images to identify morphological features and spatial relationships critical for accurate diagnoses. 
They start with a low magnification to identify suspicious regions and then zoom into image patches for detailed examination~\cite{ghezloo2022analysis, liu2024semantics}. They gather evidence across patches and accumulate them together to make a final holistic diagnosis. 
This process is the gold standard. However, it is labor-intensive and requires significant expertise to interpret complex visual information effectively. 
It is becoming increasingly unsustainable due to the rising number of cancer cases globally.

The shift towards more efficient diagnostic methods in medical imaging is essential, yet must maintain accuracy. Recent advancements in deep learning report achieving expert-level performance, promising such a scalable approach~\cite{topol2019high}. 
However, current methods typically divide WSIs into smaller patches for independent analysis, making diagnoses without the holistic context~\cite{li2021dual, yang2024foundation, zhou2024pathm3,seyfioglu2024quilt, sun2024pathgen, ahmed2024pathalign, xu2024whole, ikezogwo2023quilt, gu2023augmenting}. 
Transformer-based models attempt to capture both local and global patterns but are not scalable with the high-resolution demands of WSI~\cite{shao2021transmil, wu2021scale, guo2023higt, zheng2022graph, chen2022scaling}.

In contrast, we propose PathFinder, a multi-modal and multi-agent system designed to mimic the decision-making process of expert pathologists by integrating four AI agents: Triage, Navigation, Description, and Diagnosis. The system begins with the Triage Agent, classifying the WSI as benign or risky; if risky, the Navigation and Description Agents iteratively examine patches, generating natural language descriptions and refining their focus with each cycle. Finally, these detailed insights are integrated by the Diagnosis Agent to produce an accurate and holistic diagnostic classification. Figure \ref{fig:nav-pipeline-simple} demonstrates an overview of PathFinder's pipeline.

Our experiments demonstrate that the proposed agentic system significantly outperforms prior state-of-the-art (SOTA) methods on WSI skin melanoma grading, achieving an accuracy of 74\% on the M-Path Skin Biopsy dataset \cite{elmore2017pathologists}. This marks an 8\% improvement over the best baseline with accuracy of 66\% and a 9\% improvement over the 65\% average performance of pathologists \cite{elmore2017pathologists}. Our proposed system is also fully explainable from the patches visited to the description of the patches and the final diagnoses, which takes into consideration all the patch-wise information. To the best of our knowledge, PathFinder is the first AI-based system capable of surpassing the average performance of pathologists on this challenging melanoma classification task.

\begin{figure*}[ht!]
\centering
\includegraphics[width=\textwidth,  height=0.55\textheight]{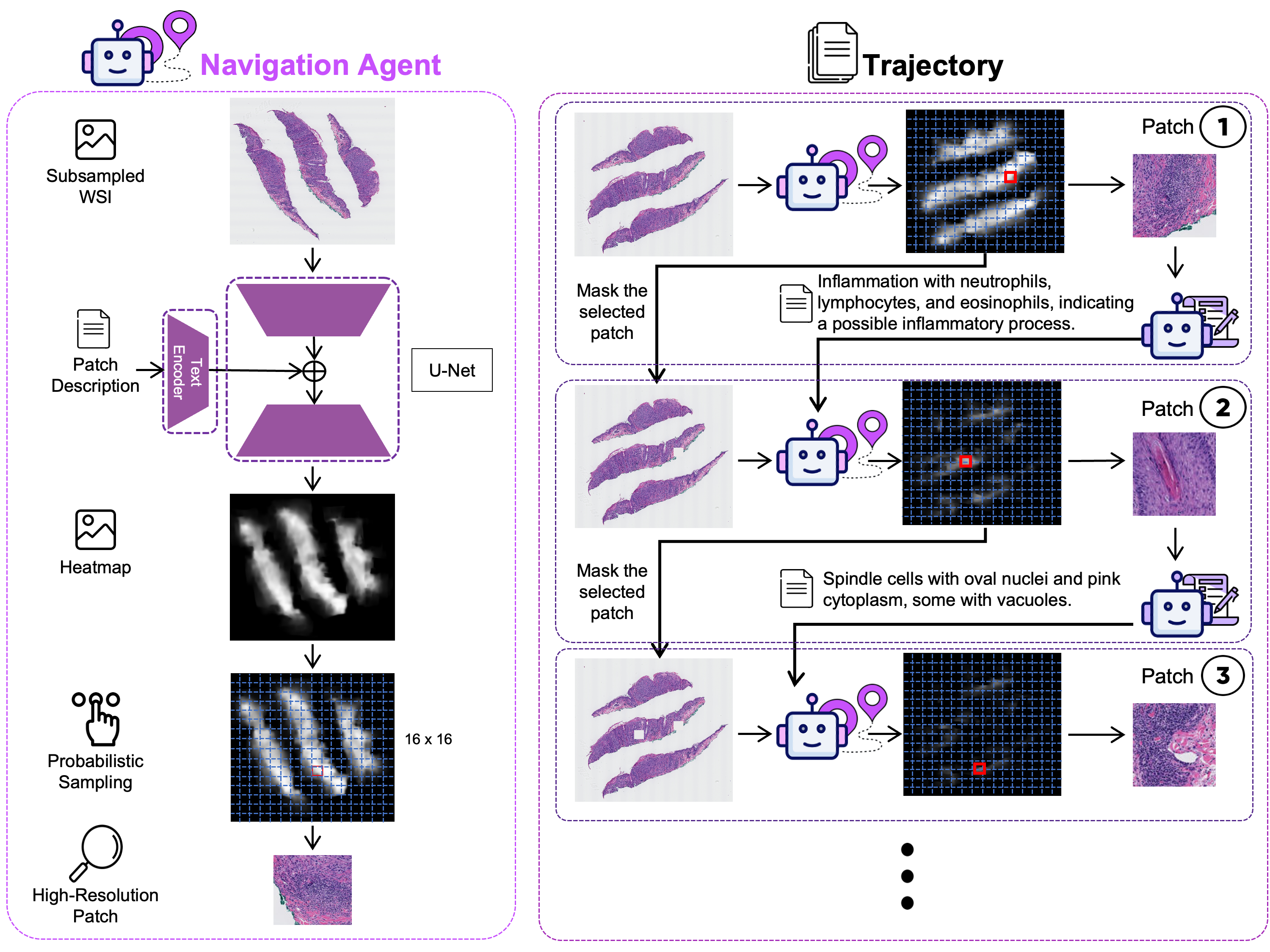}
   \caption{The left panel illustrates the Navigation Agent, as outlined in Section \ref{nav-agent}. The right panel presents the iterative trajectory generation process, which employs both the Navigation Agent and Description Agent, as described in Section \ref{trajectories}.}
\label{fig:nav-pipeline}
\end{figure*}

%% file: sec/2_related_work.tex
\section{Related Work}
\textbf{Multi-modal Histopathology Models.} 
There have been a series of studies in histopathology that leverage WSI-level and pacth-level images to train unimodal classifiers based on multiple instance modeling leveraging pretrained feature extractors \cite{shao2021transmil, scatnet, hatnet}. More recently unimodal foundational models trained on varying self-supervised objectives have achieved significant improvements on performance downstream \cite{xu2024whole, virchow, ikezogwo2022multi, uni}. With the introduction of large-scale multi-modal datasets in histopathology, we have seen significant advancements, with the emergence of large language models and vision-language models for histopathology. For instance, studies like Quilt-1M \cite{ikezogwo2023quilt} and PathGen-1.6M \cite{sun2024pathgen} curate large histopathology image-text paired dataset and train CLIP-based models to learn joint vision-language representations, significantly enhancing clinical histology downstream tasks on patch-level. On the WSI-level, PathAlign \cite{ahmed2024pathalign} aligns diagnostic texts from pathology reports with corresponding WSIs, facilitating applications such as automatic report generation and case/patient-level visual question answering, moving towards a more clinically integrated and holistic diagnostic process. While many other studies like Quilt-LLaVA \cite{seyfioglu2024quilt}, SlideChat \cite{chen2024slidechat}, and PathChat \cite{pathchat} train histopathology Multi-modal Large Language models (MLLM) and improve on diagnostic reasoning tasks, none of these models effectively automatically navigate the giga-pixel scale WSIs towards a diagnosis.

\noindent\textbf{The role of Navigation in Histopathology Diagnosis.}
Computational pathology studies have tried to capture and analyze the navigation patterns of pathologists when reviewing digital slide images \cite{roa2010experimental, mercan2018characterizing, molin2015slide, ghezloo2022analysis} specifically characterizing mouse patterns, zooming in/out, and panning the field of view (FOV) to piece out morphological clues towards a diagnosis. Often, these studies juxtapose the navigation patterns of junior and senior-level pathologists. NaviPath \cite{gu2023augmenting}, presents a human-AI collaborative navigation system designed to seamlessly integrate into pathologists' workflows, considering the specific domain knowledge and navigation strategies required for effective examination of pathology scans. 

\noindent\textbf{Multi-agent Systems.} The concept of multi-agent systems has gained traction in AI research, particularly for tasks requiring dynamic behavior and contextual understanding. Recent research has demonstrated the potential of large foundation models in creating interactive agent-based AI systems including interactions between robots, environments, and humans in the field of robotics \cite{durante2024interactive, han2024llm,wu2023autogen}. These systems can perform complex tasks by leveraging the strengths of individual agents utilizing collaboration and coordination. The potential of multi-agent systems in handling real-world scenarios has been demonstrated in recent studies including but not limited to role-playing \cite{li2303camel}, reasoning \cite{du2023improving}, gaming \cite{huang2024far} and software engineering \cite{he2024llm}. In the medical domain some studies have explored role-playing providers (clinicians) treating patients and accumulating proficiency with increasing interactions \cite{doctorsimul, medagentsbench}. These studies are centered around multiple providers; however, in medical image analysis, multi-agent systems can simulate the collaborative nature of different sub-tasks within the human-expert decision-making processes, including region navigation, understanding and holistic diagnosis.

%% file: sec/3_datasets.tex
\section{Datasets}
To lay the groundwork for describing our agents, we first start by introducing the different datasets used for training and evaluating our system.

\label{data-mpath}
\noindent\textbf{M-Path Skin Biopsy WSIs.} The skin biopsy WSIs in this dataset originate from M-Path study \cite{elmore2017pathologists, carney2016achieving, onega2018accuracy}, consisting of 238 melanocytic lesion specimens stained with Hematoxylin and eosin (H\&E). A consensus reference panel of three dermatopathologists, each with internationally recognized expertise, independently interpreted all 238 cases and established a consensus diagnosis for each case through a series of review meetings. There are 4 diagnostic classes in this dataset: class 1 with 35 cases (mild and moderate dysplastic nevi); class 2 with 86 cases (severe dysplasia/melanoma in situ); class 3 with 70 cases (invasive melanoma stage pT1a); and class 4 with 47 cases (advanced invasive melanoma stage pT1b or more). For model development, the dataset is divided into training, validation, and test sets with a 168/35/35 case split, maintaining consistent class distribution across these sets.

\noindent\textbf{M-Path Pathologists’ Viewport Data.}
The M-Path study conducted viewport data collection, recruiting 87 pathologists from 10 U.S. states. Eligibility criteria included completion of residency and/or fellowship training and recent experience interpreting skin specimens in clinical practice. Pathologists’ viewport data was gathered through an online digital slide viewer developed using Microsoft’s open-source Silverlight-based HD View SL, a gigapixel image viewer. This viewer enabled pathologists to navigate each image by panning and zooming up to 60x magnification. During interpretation, the web-based viewer automatically logged viewport tracking data, capturing a rectangular image area displayed on the pathologist's screen at any given moment. For each interpretation (unique pathologist-case pair), the system recorded a list of viewport coordinates, magnification levels, and timestamps. Data from 32 pathologists who completed the M-Path study were included in the current study. Detailed methodology of the M-Path study is available in \cite{onega2018accuracy}.

%% file: sec/3_methods.tex
\section{PathFinder}
\label{sec:methods}
The multi-agent multi-modal framework proposed in this study includes four agents: 1) Triage Agent ; 2) Navigation Agent ; 3) Description Agent ; and 4) Diagnosis Agent. The details of training data and model architectures are described below. Figure \ref{fig:nav-pipeline-simple} demonstrates how the four agents interact with each other towards the final goal which is diagnosing a WSI.

\subsection{Triage Agent}
The Triage Agent is an image-only transformer-based model tasked with separating class 1 (nevus/mild atypia and moderate atypia/dysplasia) from the rest in the M-Path dataset (Refer to section \ref{data-mpath} for M-Path class definitions). We describe the data preparation, model architecture, and training details below.

\noindent\textbf{Data Generation.} Each whole slide image (WSI) is divided into non-overlapping $512 \times 512$ patches at $10\times$ magnification. Background patches (saturation less than 15) are discarded. If fewer than 150 patches remain, we randomly select additional patches from the WSI, apply the saturation filter again, and include the ones that pass. These additional patches may overlap with existing patches but ensure that each WSI contains sufficient information. All patches are then rearranged based on their spatial coordinates. The patches are embedded using the Quilt-Net image encoder~\cite{ikezogwo2023quilt}, resulting in a feature vector of shape $(N, 768)$ per WSI, where $N$ is the total number of patches for the WSI.

\noindent
\textbf{Model Architecture.} The Triage Agent includes several sequential stages (See Fig \ref{fig:triage} in the Appendix): The feature vector is initially projected from $(N, 768)$ to $(N, dim)$ using a linear layer to align with the model’s embedding dimension $dim$. For compatibility with 2D processing, the vector is reshaped into a square grid through padding to dimensions $H \times H$, where $H$ is the smallest integer satisfying $H \times H \geq N$, with padding achieved by repeating the first $M = H^2 - N$ features. The padded vector is then processed through a transformer block, followed by positional encoding via the Pyramid Position Encoding Generator (PPEG)~\cite{shao2021transmil}, and an additional transformer block, where each transformer block contains a single self-attention layer. Subsequently, multi-scale convolutional layers and a squeeze-and-excitation (SE) block~\cite{hu2018senet} refine the vector, capturing spatial patterns across scales and emphasizing key features. The output is then flattened and transformed back to the embedding dimension. A learnable class token is appended to capture global context, and the modified vector is passed through another transformer block, positional encoding, and a final transformer block. Finally, the class token is pooled and passed through an MLP head to produce the model’s output.

\noindent
\textbf{Training Details.} We used binary cross-entropy loss for the classification task. The embedding dimension $dim$ is set to 512. Training hyperparameters are as follows: a batch size of 1, learning rate of $2 \times 10^{-4}$, weight decay of $1 \times 10^{-5}$, and gradient accumulation over 32 steps. Training is conducted for up to 100 epochs, with early stopping after 30 epochs without improvement to prevent overfitting. Our approach achieved higher F1-score and accuracy compared to other methods that are directly comparable for this task (See Table \ref{tab:triageexperiments} in the Appendix).

\subsection{Navigation Agent}
\label{nav-agent}
The Navigation Agent is designed to mimic a pathologist’s methodical approach to identifying regions of interest (ROIs) in whole slide images (WSIs). Unlike traditional systems that scan the entire WSI in a single, mechanistic sweep, our Navigation Agent adopts a more human-like, iterative process collaborating with the Description Agent. It begins by pinpointing an initial ROI, much as a pathologist would focus on one area at a time. This selected ROI is then relayed to the Description Agent, which provides a natural-language description of the area. Figure \ref{fig:nav-pipeline} illustrates the workflow of the Navigation Agent in the left panel.

In our initial attempt, we designed the Navigation Agent using a multi-modal architecture inspired by LLaVA \cite{liu2023improved}, integrating an image encoder and a large language model (LLM). The image encoder extracted features from a low-resolution version of the WSI, and the LLM processed these features along with previous text descriptions to predict the next ROI. Specifically, the WSI was divided into a grid of patches, and the LLM would output the grid coordinates of the most relevant patch based on both visual and textual inputs. However, this approach faced significant challenges due to the limited size of our training dataset. The model tended to overfit, frequently selecting central patches regardless of the input (see Appendix \ref{supp:llava-nav} for details). This limitation prompted the exploration of more data-efficient methods that could better generalize from limited samples.

To overcome these challenges, we restructured the Navigation Agent to directly generate an importance map over the WSI, conditioned on textual descriptions from previous observations. This approach removes the dependency on the LLM for spatial selection and leverages a feedback mechanism between the image and text modalities. Let $I^{(t)}$ be the input WSI at iteration $t$, with previously selected patches masked out to avoid re-sampling and $D^{(1:t)} = \{D^{(1)}, D^{(2)}, \dots, D^{(t)}\}$ be the set of textual descriptions up to iteration $t$. At each iteration $t$, the Navigation Agent processes the masked WSI $I^{(t)}$ to predict an importance map $M^{(t)}$, indicating the likelihood of each region being the next ROI. The importance map is conditioned on the aggregated textual information from previous descriptions. We define the importance map generation as $M^{(t)} = f_{\text{Nav}} \left( I^{(t)}, E^{(t-1)} \right)$ where, $f_{\text{Nav}}$ is the Navigation Agent's function (implemented as a lightweight U-Net \cite{ronneberger2015u}) that has four layers in both encoder and decoder and is conditined with text embeddings of descriptions, as well as the masked version of the WSI that masks the earlier predicted ROIs, and $E^{(t-1)}$ is the aggregated text embedding up to iteration $t - 1$. $E^{(t-1)}$ is computed by encoding each description $D^{(k)}$ using a pre-trained Text-to-Text-Transfer-Transformer (T5) text encoder \cite{raffel2020exploring} and averaging the embeddings:

\begin{equation}
    E^{(t-1)} = \frac{1}{t-1} \sum_{k=1}^{t-1} \text{T5}_{\text{text}}(D^{(k)})
\end{equation}

At the first iteration ($t = 1$), since there are no prior descriptions, the importance map is generated solely from the unmasked WSI $M^{(1)} = f_{\text{Nav}} \left( I^{(1)} \right)$.From the importance map $M^{(t)}$, we then statistically sample the next patch to analyze. The probability $p_{(i,j)}^{(t)}$ of selecting a location $(i, j)$ is proportional to its importance score:

\begin{equation}
    p_{(i,j)}^{(t)} = \frac{M_{(i,j)}^{(t)}}{\sum\limits_{(i', j')} M_{(i', j')}^{(t)}}
\end{equation}

We then sample the patch coordinates $(i^{*}, j^{*})$ based on this probability distribution:$(i^{*}, j^{*}) \sim p_{(i,j)}^{(t)}$. The selected high-resolution patch corresponding to $(i^{*}, j^{*})$ is sent to the Description Agent, which generates a new textual description $D^{(t)}$. The new description $D^{(t)}$ is encoded and incorporated into the aggregated text embedding $E^{(t)}$:

\begin{equation}
    E^{(t)} = \frac{1}{t} \sum_{k=1}^{t} \text{T5}_{\text{text}}(D^{(k)})
\end{equation}

This updated embedding $E^{(t)}$ is then used to condition the Navigation Agent in the next iteration, enabling the model to refine its importance map $M^{(t+1)}$ based on both the visual information from $I^{(t+1)}$ and the accumulated textual insights. Therefore, we refer to it as the Text-conditioned Visual Navigator.

\input{figures/ablate_fig}

\noindent
\textbf{Training details.} To train the Navigation Agent, we constructed a dataset from M-Path \cite{onega2018accuracy} consisting of WSIs and sequences of textual descriptions for the most important patches. Each training sample includes: The WSI and the corresponding masked versions, the set of descriptions $D^{(1:t)}$ for each iteration generated by Quilt-LLAVA \cite{seyfioglu2024quilt} and the ground truth importance maps derived from pathologist annotations. We minimized the binary cross-entropy loss between the predicted importance maps $M^{(t)}$ and the ground truth maps $\hat{M}^{(t)}$. Finally, to prevent overfitting, we paraphrased each description multiple times while preserving the semantic meaning, augmenting the textual data available for training.

\subsection{Description Agent}

We utilize Quilt-LLaVA \cite{seyfioglu2024quilt}, a multi-modal large language model capable of describing individual histopathology patches, as our Description Agent. While the original Quilt-LLaVA generates highly detailed findings, in this work, we instruction-tuned the model to produce more concise summaries, optimizing for computational efficiency. Using captions from the Quilt-1M dataset \cite{ikezogwo2024quilt}, we prompted GPT-4 to generate a list of findings as concise as possible. This process yielded 102,000 instruction-tuning samples. Figure \ref{fig:descriptionprompt} in the Appendix presents our prompt and a sample data entry generated for tuning the Description Agent. The Quilt-LLaVA 7B model was instruction-tuned for one epoch to obtain the Description Agent.

\subsection{Diagnosis Agent}
\label{trajectories}
The Diagnosis Agent is a language-only model that analyzes all the gathered natural text descriptions produced by the Description Agent over all the patches identified by the Navigation Agent, to analyze natural text descriptions of histopathological findings and classify them into three categories (classes 2, 3, and 4).

\noindent
\textbf{Data Generation.} To train the Diagnosis Agent, we generated diagnostic trajectories—sequences of patch descriptions that simulated how a pathologist examined a whole slide image (WSI). Using our Navigation Agent, we proceeded as follows.

We first obtained a heatmap for a sub-sampled WSI ($512 \times 512$ pixels) using a text-conditioned U-Net model, which highlighted regions of diagnostic significance. The WSI was divided into a $16 \times 16$ grid, creating 256 patches of $32 \times 32$ pixels each. Each patch received an importance score based on the mean intensity of the heatmap over the patch, indicating its diagnostic relevance. These scores were normalized across all patches.

To generate a single trajectory, we iterated the following steps ten times, yielding ten patches per case. At each iteration, a patch was selected using weighted probabilistic sampling based on the normalized importance scores, introducing variability and ensuring different patches were chosen across iterations. The selected patch was then cropped from the high-resolution $10\times$ WSI, and a description was generated by the Description Agent. Selected patches were masked on the WSI to prevent reselection, and all descriptions generated thus far were combined into a single text input for the next iteration. The generation of a single trajectory is presented in the right panel of Figure \ref{fig:nav-pipeline}.

For each WSI in the training and validation sets, we generated five ($n=5$) different trajectories, each containing ten patch descriptions, to capture various examination patterns. For the test set, we extracted additional trajectories ($n=20$) to assess the effect of trajectory number on diagnosis results. To introduce further variability, we used a LLaMA 3.1 Instruct model~\cite{dubey2024llama} at each iteration to rephrase the text descriptions. This approach effectively simulated the variability among pathologists, who might examine a single case using different patterns while seeking diagnostically relevant regions.

\noindent
\textbf{Training Details.} The Diagnosis Agent consists of a large language model (LLM) with a classification head on top. The classification head maps the LLM’s output (vocabulary size) to the number of classes, producing the final classification probabilities using a single linear layer.

We expand the training set to enhance diversity and robustness by resampling to create $20{,}000$ cases, resulting in $100{,}000$ trajectories for training. Each trajectory consists of a randomly selected number of descriptions (between five and ten), and we shuffle the sequence of descriptions within each trajectory to prevent over-fitting to any specific order. Each trajectory is formatted as a prompt to the LLM:

\textit{“The image descriptions below are extracted from different patches from the same whole slide image (WSI); please tell me which class the image belongs to: {descriptions}”},
\noindent
where \textit{{descriptions}} is the list of selected descriptions.

We fine-tune the LLM using LoRA (Low-Rank Adaptation)~\cite{hu2021loralowrankadaptationlarge} with the scaling factor $\alpha=8$, dropout rate $0.1$, and rank parameter $r=8$ in the LoRA layers. The model is trained using cross-entropy loss, with a learning rate of $5 \times 10^{-5}$, weight decay $0.001$, and batch size $16$. Due to resource constraints and the limited size of the dataset, we selected GPT-2~\cite{radford2019language} as the base pre-trained LLM.

%% file: figures/ablate_fig.tex
\begin{figure*}[ht]
    \centering
    \begin{subfigure}{0.45\textwidth}
\begin{tikzpicture}
    \begin{axis}[
        width=\linewidth,
        height=0.5\linewidth,
        xmin=0.8,xmax=20.2,
        xlabel={Num. Traj.},
        ylabel={Accuracy ($\%$)},
        grid=major,
        legend style={
            at={(0.96,0.005)}, 
            anchor=south east,
            font=\small,      
            nodes={scale=0.5, transform shape}, 
            draw=none, 
            fill=none
        },
    ]
    \pgfplotstableread[col sep=comma]{data/ntraj_vs_acc.csv}\datatable
    
    \addplot[
        thick,
        color=blue,
            mark=*,
    mark options={scale=0.8},
    ] table[x=x, y expr=\thisrow{y}  * 100] {\datatable};
    \addlegendentry{Mean Accuracy}
    
    \addplot[
        thick,
        color=white,
        name path=high,
        forget plot
    ] table[x=x, y expr=(\thisrow{y} + \thisrow{std}) * 100] {\datatable};
    
    \addplot[
        thick,
        color=white,
        name path=low,
        forget plot
    ] table[x=x, y expr=(\thisrow{y} - \thisrow{std})  * 100] {\datatable};
    
    \addplot [
        fill=orange,
        fill opacity=0.3
    ] fill between[of=high and low];
    \addlegendentry{Standard Deviation}

    \addplot [
        red,
        dashed,
        thick,
        mark=none,
    ] coordinates {(0.8,65) (20.2,65)};
\addlegendentry{Human Experts 65\%}

    \end{axis}
\end{tikzpicture}
\caption{Majority voting accuracy for 1-20 trajectories. Each trajectory contains 10 patches visited by the navigator.}
    \end{subfigure}
    \hfill
    \begin{subfigure}{0.45\textwidth}
\begin{tikzpicture}
    \begin{axis}[
        width=\linewidth,
        height=0.5\linewidth,
        xmin=0.8,xmax=10.2,
        xlabel={Traj. Length},
        ylabel={Accuracy ($\%$)},
        grid=major,
        legend style={
            at={(0.96,0.005)}, 
            anchor=south east,
            font=\small,      
            nodes={scale=0.5, transform shape}, 
            draw=none, 
            fill=none
        },
    ]
    \pgfplotstableread[col sep=comma]{data/trajLen_vs_acc.csv}\datatable
    
    \addplot[
        thick,
        color=blue,
            mark=*,
    mark options={scale=0.8},
    ] table[x=x, y expr=\thisrow{y}  * 100] {\datatable};
    \addlegendentry{Mean Accuracy}
    
    \addplot[
        thick,
        color=white,
        name path=high,
        forget plot
    ] table[x=x, y expr=(\thisrow{y} + \thisrow{std})  * 100] {\datatable};
    
    \addplot[
        thick,
        color=white,
        name path=low,
        forget plot
    ] table[x=x, y expr=(\thisrow{y} - \thisrow{std}) * 100] {\datatable};
    
    \addplot [
        fill=orange,
        fill opacity=0.3
    ] fill between[of=high and low];
    \addlegendentry{Standard Deviation}

    \addplot [
        red,
        dashed,
        thick,
        mark=none,
    ] coordinates {(0.8,65) (20.2,65)};
    \addlegendentry{Human Experts 65\%}

    \end{axis}
\end{tikzpicture}
\caption{Majority voting accuracy for 5 trajectories with 1-10 patches per trajectory.}
    \end{subfigure}
    \caption{Ablation results. We ran 10 experiments, and plotted both the mean and standard deviation.}
    \label{fig:trajectory-analysis}
\end{figure*}
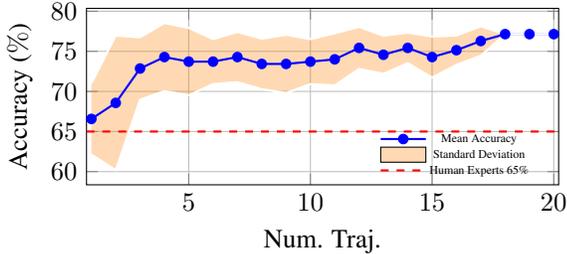
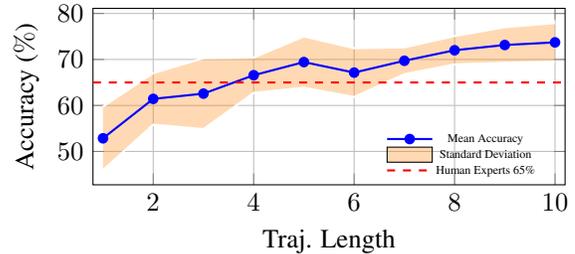

%% file: sec/4_Experiments.tex
\section{Experiments and Results}

This section outlines the experimental setup and evaluates the performance of the proposed PathFinder framework. First, we conduct a qualitative assessment of the descriptions generated by the Description Agent, comparing them to two vision-language models (VLMs). Next, we evaluate PathFinder on the M-Path dataset for melanoma diagnosis (see Section \ref{data-mpath}), benchmarking it against state-of-the-art transformer-based and MIL-based baselines, as well as public and private large language models (LLMs) using prompting without additional training. Finally, we analyze PathFinder's performance under various configurations, altering the Triage, Navigation, and Description Agents. Detailed evaluations are provided in the following subsections.

\subsection{Pathologist Evaluation of Description Quality}
\label{sup:pathologist-analysis}

To assess the quality of descriptions generated by our Description Agent, we conducted a survey in which two expert pathologists rated descriptions produced by our Description agent in comparison to those generated by GPT4-o \cite{hurst2024gpt} and LLaVA-Med \cite{li2024llava}. We selected 25 cases from the M-Path dataset, sampling across the four diagnostic classes. For each case, we cropped the consensus region of interest, manually labeled by a panel of expert dermatopathologists as the area most representative of the diagnosis. Using this region, we prompted our Description Agent, LLaVA-Med and GPT4-o to generate concise descriptions of each histopathology patch. These descriptions were then presented to two expert pathologists in a randomized, double-blind format. Each pathologist was asked to respond to two questions for each case to indicate their preferred description and the reason for their preference. Refer to Appendix \ref{supp:pathologists-analysis} for a detailed explanation of this assessment. The results shown in Figure \ref{fig:pathologist-analysis} indicate that, PathFinder's Description agent achieves comparable performance to GPT-4o while being significantly more cost-effective, operating with just 7B parameters - a fraction of GPT-4o's size.

\input{figures/preference}

\subsection{PathFinder Evaluation}

For evaluating PathFinder, we utilize the M-Path dataset which contains histopathology WSIs of melanocytic skin tissue. As outlined in \ref{trajectories}, multiple trajectories are generated per case to simulate the variability in diagnostic patterns observed among pathologists, who may assess a single case with diverse visual strategies to identify diagnostically significant regions. To mitigate randomness in our results, we evaluated PathFinder 10 times on the test set, each time using a different random subset of 5 trajectories selected from the total of 20. For each Whole Slide Image (WSI), majority voting is performed on the predictions from the 5 selected trajectories to produce the final result. The overall performance is then reported in Table \ref{tab:navigationdiagnosis} as the mean of the results across the 10 runs. We balanced the testing dataset to ensure that each diagnostic class is represented by an equal number of samples. Consequently, the micro-averaged F1 score, precision, and recall are equivalent to the accuracy reported in the table. We opted to use micro-averaged metrics in our clinical evaluation, because they appropriately balance the importance of different stages of skin cancer, which is crucial for assessing the overall reliability and effectiveness of the diagnostic tool.

\begin{table*}[h]
  \centering
  \small
  \begin{tabular}{lcc}
    \toprule
    Methods & Accuracy & F-1 score \\ 
    \midrule
    \textit{Baselines} &  & \\
    \hdashline
    Human Experts \cite{elmore2017pathologists} & 0.65  & 0.65 \\
    \hdashline
    ScAtNet \cite{wu2021scale} & 0.62 & 0.62 \\ 
    ScAtNet + ROI Heatmap \cite{ghezloo2024} & 0.63 & 0.63 \\ 
    ScAtNet + SAG \cite{liu2024semantics} & 0.60 & 0.60 \\  
    ABMIL \cite{ilse2018attention}* & 0.46 & 0.47 \\ 
    ABMIL w/ CONCH \cite{conch}* & 0.66 & 0.60 \\
    ABMIL w/ UNI2-h \cite{uni}* & 0.66 & 0.66 \\
    ABMIL w/ QuiltNet \cite{ikezogwo2023quilt}* & 0.61 & 0.63 \\
    \midrule
    \textit{LLM Prompting Baselines} &  & \\
    \hdashline
    BioMistral-7B & 0.43 & 0.43 \\
    Mistral-Nemo-Instruct-2407 & 0.41 & 0.41 \\
    GPT-4o & 0.49 & 0.49 \\
    Meta-Llama-3-8B-Instruct & 0.31 & 0.31 \\
    LLaVA-Med-v1.5-Mistral-7b & 0.43 & 0.43 \\
    Quilt-LLaVA-v1.5-7b & 0.29 & 0.29 \\
    \midrule
    \textit{Ours} &  & \\
    \hdashline
    PathFinder + ABMIL w/ UNI2-h Attention-Based Top Patches & 0.46 & 0.46 \\
    PathFinder + ABMIL w/ QuiltNet Attention-Based Top Patches & 0.46 & 0.46 \\
    PathFinder + ABMIL w/ CONCH Attention-Based Top Patches & 0.54 & 0.54 \\
    PathFinder + T5-Based Text-Conditioned Visual Navigator + No Triage Agent & 0.58 & 0.58 \\
    PathFinder + T5-Based Text-Conditioned Visual Navigator + LLaVA-Med Descriptions & 0.56 & 0.56 \\
    PathFinder + CLIP-Based Text-Conditioned Visual Navigator + LLaVA-Med Descriptions & 0.60 & 0.60 \\
    PathFinder + Imitated Sampling & 0.63 & 0.63 \\
    PathFinder + Vision-Only Navigator & 0.64 & 0.64 \\
    PathFinder + CLIP-Based Text-Conditioned Visual Navigator & 0.62 & 0.62 \\
    PathFinder + Exhaustive search & 0.68 & 0.68 \\
    \textbf{PathFinder + T5-Based Text-Conditioned Visual Navigator} & \textbf{0.74} & \textbf{0.74} \\
    \bottomrule
    \vspace{1em}
  \tiny{* ABMIL result is based on a single run and does not use majority voting}
  \end{tabular}
  \caption{Majority voting performance for whole slide image (WSI) diagnosis on the M-Path dataset. Accuracy is reported, and the F-1 score is identical due to the balanced testing set. Finally, coverage here is the percent of patches used across all trajectories.}
  \label{tab:navigationdiagnosis}
\end{table*}

We compared Pathfinder to four state-of-the-art baseline models: 1) three transformer-based models all utilizing the ScAtNet architecture \cite{wu2021scale, ghezloo2024, liu2024semantics} and 2) four MIL-based model, using ABMIL \cite{ilse2018attention} with different backbones. ScAtNet utilizes a MobileNetV2 backbone \cite{sandler2018mobilenetv2} to extract multi-scale features from images at 7.5x, 10x, and 12.5x magnification. For the first baseline model, these feature vectors are subsequently fed into ScATNet which aggregates information of the three scales to perform the diagnostic task using Transformer blocks. The second approach \cite{ghezloo2024} augments the WSI with ROI heatmaps generated by the U-Net model, appending these maps as a fourth input channel and using ScAtNet for classification. The third baseline model, SAG \cite{liu2024semantics}, converts diagnostically relevant entities into attention signals, integrating these with ScAtNet and employing an attention-guiding loss function that combines heuristic guidance (HG) and tissue guidance (TG) based on disease-specific prior knowledge such as tissue, structure, and cellular information. In addition to the original ABMIL model \cite{ilse2018attention}, we extended our evaluation by incorporating three additional ABMIL variants, each using a different pathology-specific foundation model as a backbone: CONCH \cite{conch}, UNI2-h \cite{uni}, and QuiltNet \cite{ikezogwo2023quilt}. ABMIL aggregates information across instances using an attention mechanism that assigns weights to each instance, allowing the model to capture its contribution to the final bag label in a permutation-invariant manner.

Then, we conducted comprehensive experiments to evaluate PathFinder by examining different architectures for each agent component, achieving 74\% accuracy that surpasses both human experts (65\%) and previous state of the art (66\% best). Our evaluation focused on three main aspects:

\noindent
\textbf{Navigator Architectures.} First, to quantify the importance of Description Agent feedback, we tested a Visual-Only Navigator that employs weighted probabilistic sampling for patch selection without iterative feedback in a single pass over the WSI. Additionally, we implemented Imitated Sampling, which leverages pathologists' viewing pattern distributions (viewport width, height, and zoom level) from our M-Path dataset (Section \ref{data-mpath}) to statistically sample patches as important WSI regions. If pathologists spend more time focusing on a region, we gave a higher chance of sampling to that region.  Both Imitated Sampling and Vision-Only Navigator performed similarly (64\% and 63\% respectively), indicating that both pure statistical and learned "sampling", regardless of source, has limited effectiveness. 
To further assess the necessity of our iterative navigation approach, we added a non-iterative baseline, selecting the top 10 patches from ABMIL attention scores using three different pathology-specific foundation model backbones. The best model with CONCH\cite{conch} backbone performed substantially lower than our best navigation-based approach (74\% vs 54\%). This confirms that a purely image-based, non-iterative selection approach is insufficient, as it lacks the ability to iteratively refine patch selection based on evolving textual descriptions. 
Furthermore, we evaluated text-conditioned visual navigators using either CLIP-based or T5-based text encoders. The T5-based navigator significantly outperformed its CLIP-based counterpart (74\% vs 62\%), suggesting CLIP's 77-token limit constrains its ability to effectively process multiple descriptions (we simply truncate descriptions exceeding 77 tokens, then average if a description is long). Finally, our navigation-based approach (74\%) outperformed exhaustive search (68\%), which utilizes all non-background patches of the WSI, suggesting that selective patch sampling helps avoid confusion from irrelevant regions.

\noindent
\textbf{Description Agents.} We compared a fine-tuned version of Quilt-LLaVA (optimized for concise descriptions) against off-the-shelf LLaVA-Med. The fine-tuned version showed superior performance (74\% vs 56\%), demonstrating better guidance for the navigator. Notably, when paired with LLaVA-Med descriptions, the T5-based Navigator showed no advantage over the CLIP-based version (56\% vs 60\%). This suggests that a more powerful text encoder like T5 can actually be detrimental when processing lower-quality descriptions, potentially steering the Navigator toward irrelevant regions. This finding emphasizes the importance of high-quality descriptions for effective navigation.

\noindent
\textbf{Diagnosis Agents.} We evaluated various public and private LLMs as baselines for our Diagnosis Agent(detailed in Table \ref{tab:navigationdiagnosis} under LLM Prompting Baselines). Specifically, we used PathFinder with T5-Based Text-Conditioned Visual Navigator and Quilt-LLaVA Description Agent to generate multiple descriptions for each WSI and prompted LLMs to make the classification given the descriptions. To view our prompt, please see Section \ref{supp:diagnosis-llm-prompting} in Appendix.

Finally, it is worth noting that without the Triage Agent, the performance of the best Pathfinder-variant dropped below baselines, likely due to Quilt-LLaVA's train dataset's bias toward malignant cases.

The evaluation of the baseline models are similarly done using the majority voting over 10 runs. PathFinder achieves 8\% improvement compared to the best baseline approaches, ABMIL with CONCH \cite{conch} and UNI2-h \cite{uni} backbones. Considering that GPT-2 is a relatively small LLM compared to the current state-of-the-art, we believe that utilizing larger LLMs could further improve diagnostic outcomes.

Lastly, to investigate the impact of the number of trajectories on model performance, we evaluated the model using between 1 and 20 trajectories for majority voting, as well as the effect of varying trajectory lengths. Figure \ref{fig:trajectory-analysis} illustrates this analysis, indicating optimal performance with 5 trajectories and 10 patches per trajectory. We run every experiment for 10 rounds and report the mean and standard deviation.

%% file: figures/preference.tex
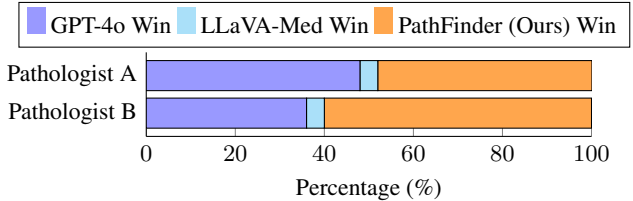
\begin{figure}[h]
    \centering
\begin{tikzpicture}
    \begin{axis}[
        width=0.9\linewidth,
        height=2.7cm,
        xbar stacked,
        bar width=0.4cm, 
        xmin=0, xmax=100,
        axis x line*=bottom,
        axis y line*=left,
        ymin=-0.2, ymax=1.2, 
        ytick=data,
        yticklabels={Pathologist B, Pathologist A},
        enlarge y limits=0.3,
        xlabel={Percentage (\%)},
        tick label style={font=\small}, 
        label style={font=\small},      
        legend style={font=\small, at={(0.4,1.6)}, anchor=north, legend columns=-1},
        legend image code/.code={
            \draw[#1, draw=none] (0cm,-0.1cm) rectangle (0.25cm,0.25cm);
        },
    ]
    \addplot[fill=blue!40] coordinates {(36,0) (48,1)}
        node[anchor=east, xshift=-1em] at (axis cs:0,0) {Pathologist A's score};  
    
    \addplot[fill=cyan!30] coordinates {(4.0,0) (4.0,1)}
        node[anchor=east, xshift=-1em] at (axis cs:0,1) {Pathologist B's score};  
    
    \addplot[fill=orange!70] coordinates {(60,0) (48.1,1)};
    
    \legend{GPT-4o Win, LLaVA-Med Win, PathFinder (Ours) Win}
    \end{axis}
\end{tikzpicture}

\caption{Expert human pathologist preferences for each model in assessing description quality, evaluated in a double-blind survey for unbiased comparison.}

\label{fig:pathologist-analysis}
\end{figure}

%% file: sec/5_Discussion.tex
\section{Discussion}

This study presents PathFinder, a multi-modal, multi-agent AI framework designed to emulate the multi-scale, iterative diagnostic approach of expert pathologists for histopathology whole slide images (WSIs). By integrating Triage, Navigation, Description, and Diagnosis Agents, PathFinder collaboratively gathers evidence to deliver accurate, interpretable diagnoses with natural language explanations. Notably, it surpasses state-of-the-art methods and the average performance of human experts in melanoma diagnosis, setting a new benchmark in AI-driven pathology.

PathFinder has the potential to accelerate diagnostic workflows, reducing the reliance on manual examination and enabling timely patient care in clinical settings. Its natural language descriptions provide interpretability, facilitating the validation of AI-generated diagnoses by pathologists. Moreover, its integration of vision-language models (VLMs) and large language models (LLMs) highlights the promise of multi-modal AI in delivering scalable, specialized diagnostic tools that could improve access to pathology expertise.

\noindent\textbf{Limitations.} Despite its strengths, PathFinder has limitations. The framework relies on pre-existing datasets and significant computational resources, posing challenges in resource-constrained environments. Additionally, the complexity of the Navigation Agent’s decision-making process and occasional hallucinations by the Description Agent could affect transparency and accuracy of the decision-making process. Future work should address these issues by enhancing dataset diversity, computational efficiency, and patch selection strategies, further advancing PathFinder's potential as a transformative tool in AI-assisted pathology.

%% file: sec/X_suppl.tex
\clearpage
\setcounter{page}{1}
\setcounter{section}{0}  
\setcounter{figure}{0}   
\setcounter{table}{0}    
\setcounter{equation}{0} 
\maketitlesupplementary

\section{Triage agent}

\begin{figure*}[ht]
\begin{center}
\includegraphics[width=\linewidth]{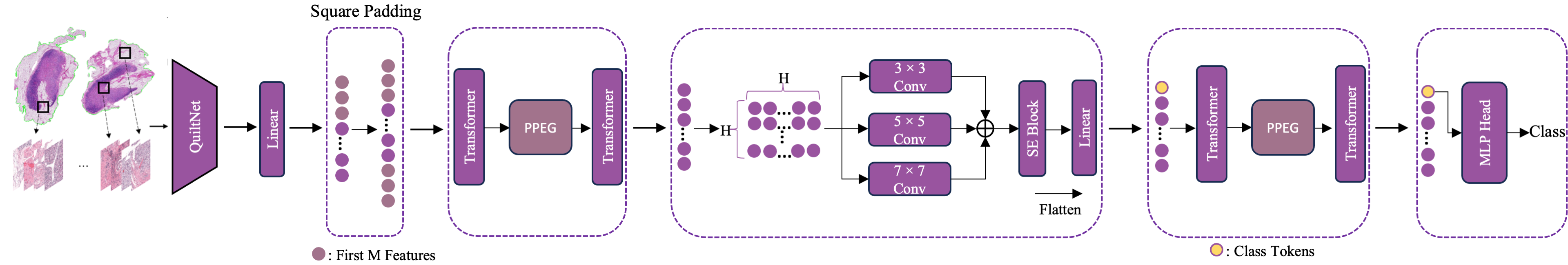}
\end{center}
   \caption{Overview of the Triage Agent architecture. Definitions of $M$ and $H$ can be found in Section \ref{sec:methods}.}
\label{fig:triage}
\end{figure*}

Figure \ref{fig:triage} illustrates the architecture of the Triage Agent. To evaluate its effectiveness, we compared the performance of the Triage Agent against three MIL-based benchmark methods \cite{ilse2018attentionbaseddeepmultipleinstance, li2021dualstreammultipleinstancelearning, shao2021transmil} for detecting Class 1 vs. Non-Class 1 cases in the M-Path dataset (details in Section \ref{data-mpath}). As summarized in Table \ref{tab:triageexperiments}, PathFinder's Triage Agent, designed to assess whether a WSI is risky, outperforms the baseline methods.

\begin{table}[ht]
  \centering
  \scriptsize
  \begin{tabular}{lccc}
    \toprule
    Method & Class 1 F1 & Non-Class 1 F1 & Overall Accuracy \\
    \midrule
    AMIL \cite{ilse2018attentionbaseddeepmultipleinstance} & 0.16 & 0.83 & 0.71 \\
    DSMIL \cite{li2021dualstreammultipleinstancelearning} & 0.35 & 0.86 & 0.77 \\
    TransMIL \cite{shao2021transmil} & 0.40 & 0.90 & 0.83 \\
    \midrule
    \textbf{Triage Agent} & \textbf{0.57} & \textbf{0.95} & \textbf{0.91} \\
    \bottomrule
  \end{tabular}
  \caption{Comparison of Triage Agent with benchmark methods on Class 1 vs. Non-Class 1 classification}
  \label{tab:triageexperiments}
\end{table}

\section{VLM-based Navigation Agent}
\label{supp:llava-nav}

Our initial approach to designing the Navigator Agent explored a multi-modal architecture based on the LLaVA framework \cite{liu2023improved}. This design aimed to enable direct reasoning over image latents through an LLM. The architecture consisted of two main components:

\begin{enumerate}
    \item A U-Net encoder \cite{ronneberger2015u} pre-trained on pathologist viewing behavior data (M-Path, details in Section \ref{data-mpath}), which served as the image encoder
    \item The LLaMA-7B language model \cite{touvron2023llama}, which acted as the reasoning component
\end{enumerate}

\subsection{Training Process and Architecture}
We first trained a complete U-Net on the M-Path dataset to learn meaningful representations of WSIs. For the Navigator implementation, we removed the U-Net's decoder and retained only the encoder portion. This encoder was then integrated with LLaMA-7B following the LLaVA framework. The combined model was trained using instruction tuning, where each training instance consisted of:
\begin{itemize}
    \item Input: A WSI and a list of previous observations and their descriptions obtained from the Description agent
    \item Output: Grid coordinates (row and column) identifying regions of interest within the WSI
\end{itemize}

The underlying hypothesis was that the LLM could effectively process the U-Net-encoded latent representations to identify diagnostically relevant grid coordinates directly.

\subsection{Limitations and Challenges with a LLaVA-based Navigator}
This approach encountered several significant limitations:

\begin{enumerate}
    \item \textbf{Data Scarcity}: The available navigation training dataset proved insufficient for the model to learn robust region selection strategies.
    
    \item \textbf{Overfitting Patterns}: The model exhibited clear signs of overfitting:
    \begin{itemize}
        \item Consistently selecting patches from the central regions of WSIs, regardless of input
        \item Generating repetitive patch selections
        \item Failing to generalize to novel slide patterns
    \end{itemize}
\end{enumerate}

\subsection{Architectural Pivot}
These limitations led us to revise our approach fundamentally. Instead of requiring the LLM to reason directly from latent representations, we returned to utilizing the complete U-Net architecture (including the decoder), and leverage the decoded attention maps for direct region sampling. This proved to be more robust with limited training data, and we simply conditioned our U-Net with the descriptions from the Description Agent to have the feedback loop between the agents. This experience highlighted the challenges of applying LLMs to specialized medical tasks with constrained training data, even when pre-training sub-modules (like our U-Net encoder in this case).

\section{Description agent}

\begin{figure*}[h]
\begin{center}
\includegraphics[width=0.9\linewidth]{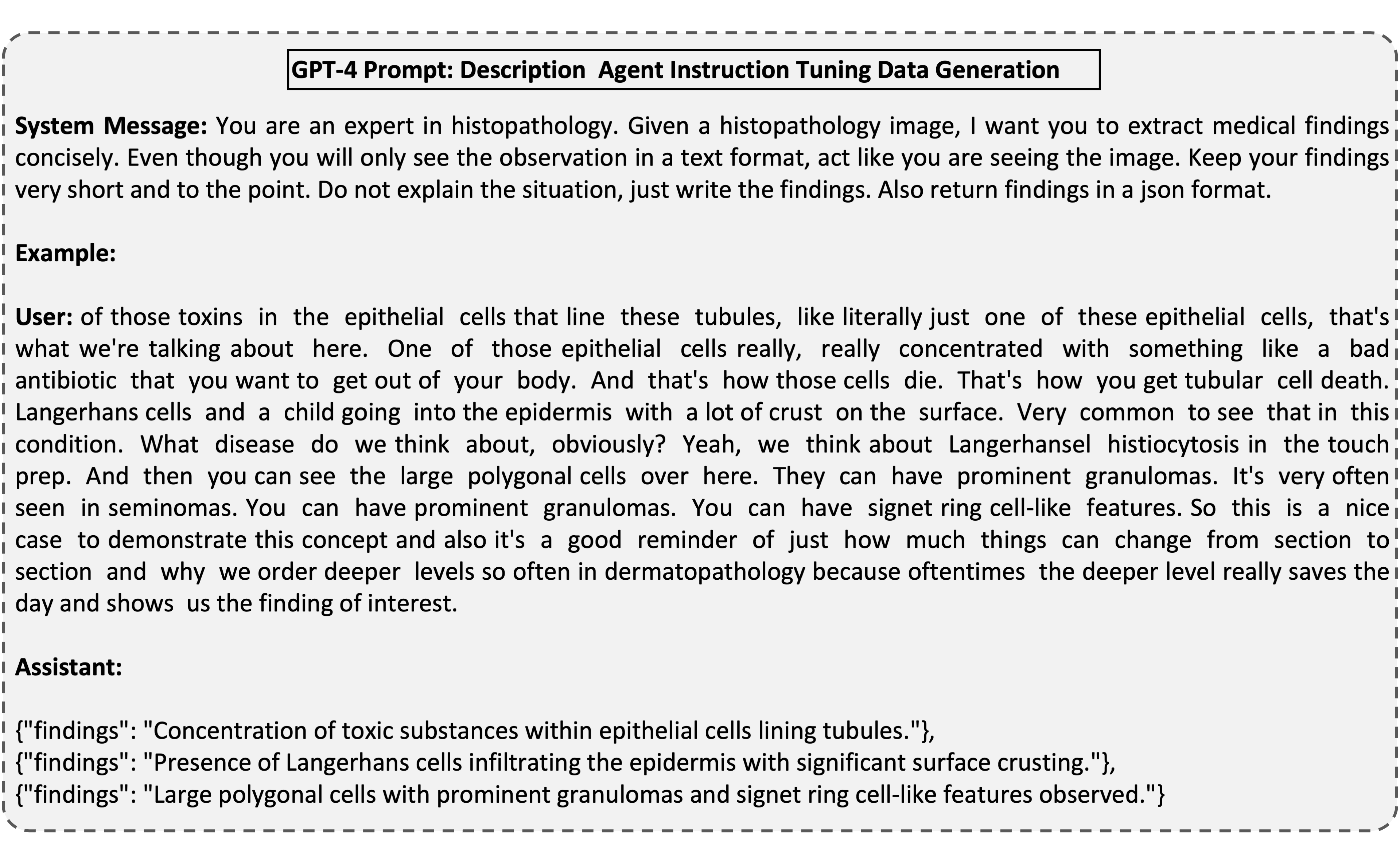}
\end{center}
   \caption{GPT-4 prompt to generate instruction-tuning dataset for the Description Agent.}
\label{fig:descriptionprompt}
\end{figure*}

We generated fine-tuning data for the Description Agent by prompting GPT-4 to extract short and concise histopathology findings from provided text. Figure \ref{fig:descriptionprompt} illustrates the prompt used and a sample of the data generated for fine-tuning the Description Agent.

\section{Evaluation and experiments}

This section provides details on the qualitative analysis conducted by pathologists and the prompt for our LLM-prompting experiments.

\subsection{Qualitative Analysis of Descriptions Assessed by Pathologists}
\label{supp:pathologists-analysis}

To evaluate the quality of the descriptions generated by the Description Agent, we cropped the region of interest from 25 WSIs from M-Path dataset and generated descriptions for these regions using three models: PathFinder's Description Agent, GPT-4o, and LLaVa-Med. Since our Description Agent is fine-tuned to produce short and concise descriptions, we ensured a fair comparison by prompting LLaVa-Med and GPT-4o with the instruction: \textit{Describe the histology image concisely in less than 20 words}. We conducted a survey involving two pathologists who were asked to answer the following two questions regarding descriptions produced by the three models. The study was conducted in a double-blind, randomized manner to ensure unbiased results:

\begin{enumerate}
    \item \textbf{Selection}: Please select the description that you believe best matches the content of the image. (Options: Model A, Model B, Model C)
    \item \textbf{Reason for Preference}: Please choose the primary reason for your preference. You may select more than one option if applicable. If Other, please specify.
    \begin{itemize}
        \item \textbf{Correctness}: The description accurately reflects the features of the image.
        \item \textbf{Detail}: The description provides a comprehensive analysis of the image.
        \item \textbf{Relevance}: The description emphasizes the most pertinent aspects of the image.
        \item \textbf{Other}: Please specify.
    \end{itemize}
\end{enumerate}

\input{figures/preference-reason}

Figure \ref{fig:pathologist-analysis-reason} illustrates the distribution of reasons selected by pathologists for preferring each model. As shown, none of the models were preferred for their level of detail, which aligns with expectations since the models were specifically prompted to generate short and concise descriptions, inherently limiting detailed information. The majority of preferences were based on the correctness of the descriptions.

\subsection{Prompt used for pre-trained LLM experiments}
\label{supp:diagnosis-llm-prompting}
The following prompt was used in our experiments with pre-trained LLMs serving as the Diagnosis Agent to make a diagnosis based on the provided \textit{descriptions}:

\noindent\textbf{Prompt:} Answer the following question related to skin cancer. Only use one of the four options given at the end. \newline
The image descriptions below are extracted from different patches from the same whole slide image (WSI), please tell me which class the image belongs to:\newline
\textit{\{descriptions\}}\newline
The options are: \\
"diagnosis: (I) mildly dysplastic nevi, moderately dysplastic nevi"\\
"diagnosis: (II) melanoma in situ and severely dysplastic nevi"\\
"diagnosis: (III) invasive melanoma stage pT1a"\\
"diagnosis: (IV) advanced invasive melanoma stage $\geq$ pT1b"\\
Only output the complete text of the option you choose. Don't add any more words.

%% file: figures/preference-reason.tex
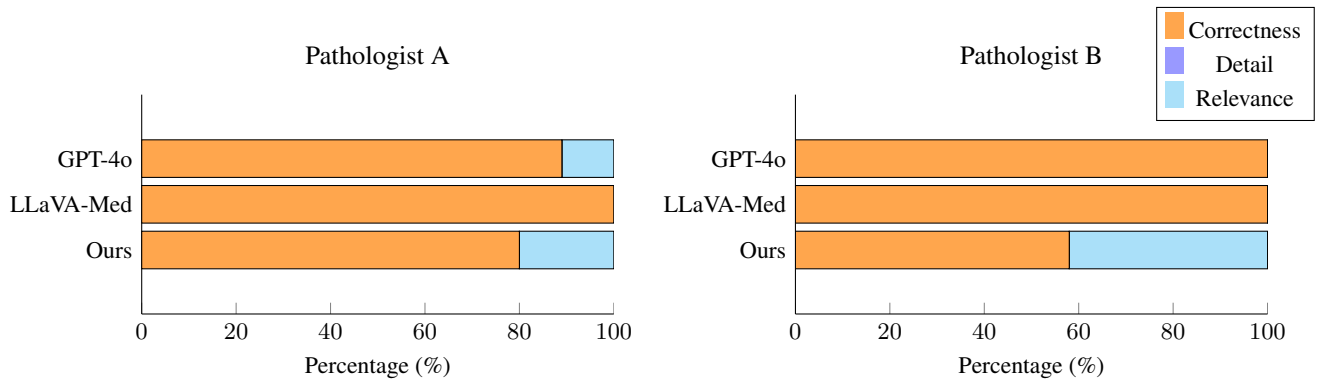
\begin{figure*}[h]
    \centering
\begin{tikzpicture}
    \begin{axis}[
        width=0.45\linewidth,
        height=4.5cm,
        xbar stacked,
        bar width=0.5cm,
        xmin=0, xmax=100,
        axis x line*=bottom,
        axis y line*=left,
        ymin=-0.5, ymax=2.5,
        ytick=data,
        yticklabels={Ours, LLaVA-Med, GPT-4o},
        enlarge y limits=0.3,
        xlabel={Percentage (\%)},
        title={Pathologist A},
        tick label style={font=\small},
        label style={font=\small},
    ]
    \addplot[fill=orange!70] coordinates {(80,0) (100,1) (89,2)}; 
    \addplot[fill=blue!40] coordinates {(0,0) (0,1) (0,2)}; 
    \addplot[fill=cyan!30] coordinates {(20,0) (0,1) (11,2)}; 
    
    \end{axis}
\end{tikzpicture}
\hfill
\begin{tikzpicture}
    \begin{axis}[
        width=0.45\linewidth,
        height=4.5cm,
        xbar stacked,
        bar width=0.5cm,
        xmin=0, xmax=100,
        axis x line*=bottom,
        axis y line*=left,
        ymin=-0.5, ymax=2.5,
        ytick=data,
        yticklabels={Ours, LLaVA-Med, GPT-4o},
        enlarge y limits=0.3,
        xlabel={Percentage (\%)},
        title={Pathologist B},
        tick label style={font=\small},
        label style={font=\small},
        legend style={font=\small, at={(1.1,1.4)}, anchor=north east, legend columns=1},
        legend image code/.code={
            \draw[#1, draw=none] (0cm,-0.1cm) rectangle (0.25cm,0.25cm);
        },
    ]
    \addplot[fill=orange!70] coordinates {(58,0) (100,1) (100,2)}; 
    \addplot[fill=blue!40] coordinates {(0,0) (50,1) (40,2)}; 
    \addplot[fill=cyan!30] coordinates {(42,0) (20,1) (30,2)}; 
    
    \legend{Correctness, Detail, Relevance}
    \end{axis}
\end{tikzpicture}

\caption{Expert human pathologist preferences for each model, segmented by the reasons for their choices. Each subplot corresponds to one pathologist and shows their ratings for PathFinder (Ours), LLaVA-Med, and GPT-4o.}
\label{fig:pathologist-analysis-reason}
\end{figure*}